\documentclass[11pt]{article}

\usepackage[final]{acl}

\usepackage{times}
\usepackage{latexsym}

\usepackage[T1]{fontenc}
\usepackage[utf8]{inputenc}

\usepackage{microtype}

\usepackage{inconsolata}

\usepackage{graphicx}

\usepackage{wrapfig}
\usepackage{xspace}
\usepackage[most]{tcolorbox}
\usepackage[dvipsnames]{xcolor}
\usepackage[framemethod=TikZ]{mdframed}
\usepackage{amsmath}
\usepackage{algorithm}
\usepackage{algorithmic}
\usepackage{booktabs}

\usepackage{booktabs}
\usepackage{graphicx}
\usepackage{pifont}
\usepackage[table,xcdraw]{xcolor}
\usepackage[table]{xcolor}
\definecolor{retainGreen}{RGB}{235,245,235}  % pale green
\definecolor{oursRed}{RGB}{252,235,235}  

\newcommand{\methodname}{\texttt{CAU}\xspace}

\usepackage{soul}% for underlines

\usepackage{amsmath}
\usepackage{amssymb}
\usepackage{mathtools}
\usepackage{amsthm}
\usepackage{graphicx}
\usepackage{subcaption}

\definecolor{darkgreen}{RGB}{0,80,60}

\theoremstyle{plain}

\theoremstyle{definition}

\theoremstyle{remark}

\usepackage[textsize=tiny]{todonotes}
\usepackage{lineno}

\definecolor{darkblue}{rgb}{0, 0, 0.5}
\hypersetup{colorlinks=true, citecolor=darkblue, linkcolor=darkblue, urlcolor=darkblue}

\usepackage[capitalize,noabbrev]{cleveref}
\title{When Retain Constraints Conflict: Mitigating Forget–Retain Interference in Tabular Data}

\author{
  \textbf{Zijie Liu}$^{1}$ \quad
  \textbf{Jinhao Duan}$^{1}$ \quad
  \textbf{Bingqi Shang}$^{2}$ \quad
  \textbf{Xinming An}$^{1}$ \\
  \textbf{Sijia Liu}$^{2}$ \quad
  \textbf{Tianlong Chen}$^{1,\dagger}$ \\
  $^{1}$University of North Carolina at Chapel Hill \\
  $^{2}$Michigan State University \\
  $^{\dagger}$Corresponding author: \texttt{tianlong@cs.unc.edu}
}
\begin{document}
\maketitle
\begin{abstract}

% 1. though unlearnign is important, yet it is underexpored on tabular data
% 2. the key challenge that differentiate from other modality/use case, text/vlm/xxx, is the overlap
% 3. explain in detail
% 4. motivated by that?, we propose xxx (brief intro)
% 5. extend with exp results.

Machine unlearning aims to remove the influence of designated training data while preserving model utility, but its behavior on tabular data remains underexplored. This gap is important because tabular prediction is widely used in high-stakes domains and is increasingly adapted to language models through record serialization and schema-aware prompting. We identify a key challenge that distinguishes tabular unlearning from unlearning in free-form text or other modalities: schema-induced forget–retain overlap. In serialized tabular data, records share fixed column-name/value slots, similar attribute ranges, and common output spaces. Consequently, a forget row may have nearby retain rows that rely on the same high-signal attributes, causing retain preservation to oppose the update required for forgetting. Motivated by this failure mode, we propose Conflict-Aware Unlearning (CAU), a schema-aware approach that reduces forget–retain interference by relaxing preservation constraints on retained rows that most conflict with the forget set. Across sample-level and feature-level unlearning on clinical and non-medical tabular tasks, \methodname more closely matches a retraining oracle while maintaining predictive utility and retain-region behavior. Our results show that reliable tabular LLM unlearning depends not only on the forgetting objective, but also on how retain constraints are constructed.
\end{abstract}

\section{Introduction}
\label{sec:intro}
\begin{figure}[t]
    \centering

    \begin{subfigure}[t]{0.48\linewidth}
        \centering
        \includegraphics[width=\linewidth]{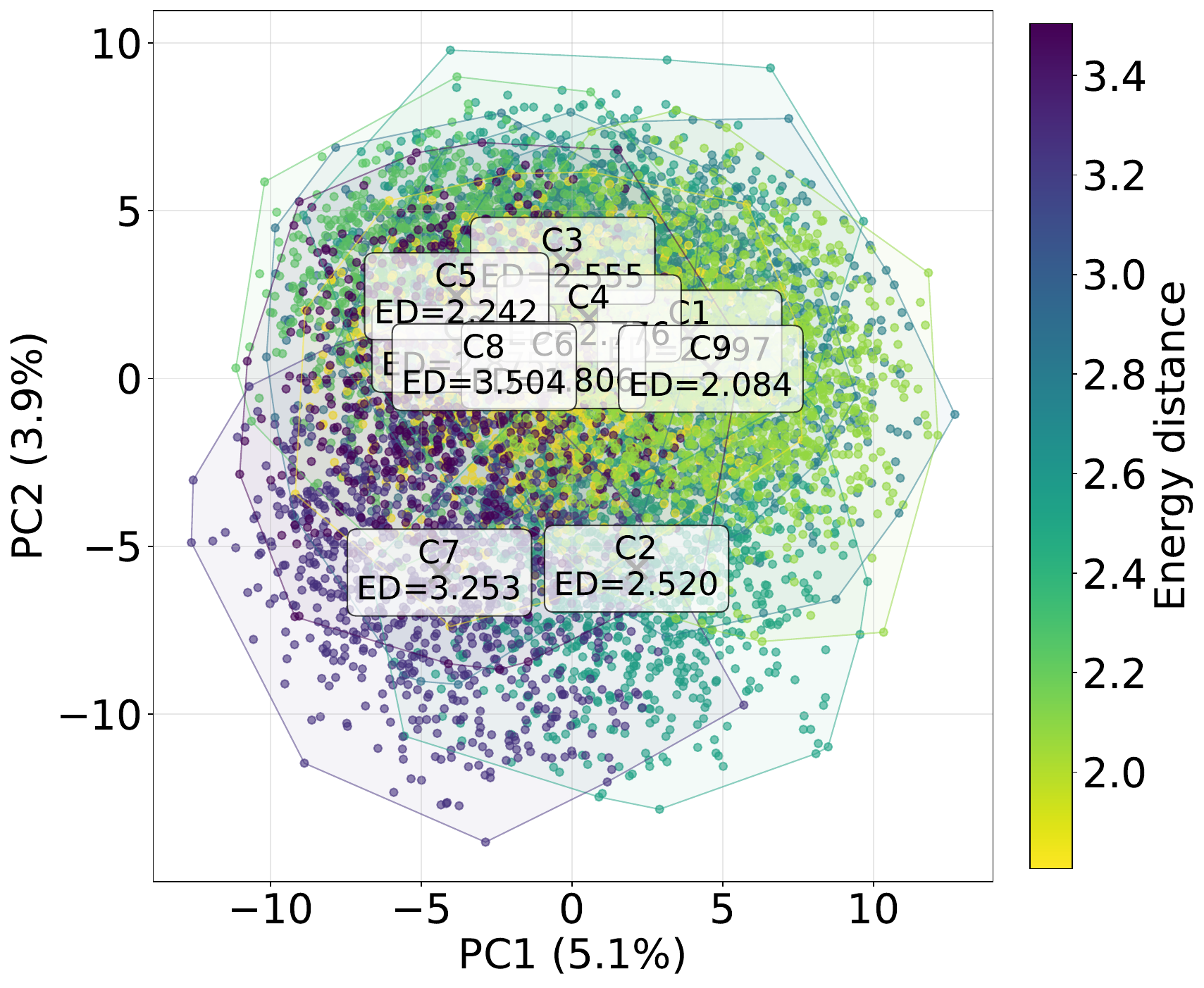}
    \end{subfigure}
    \hfill
    \begin{subfigure}[t]{0.48\linewidth}
        \centering
        \includegraphics[width=\linewidth]{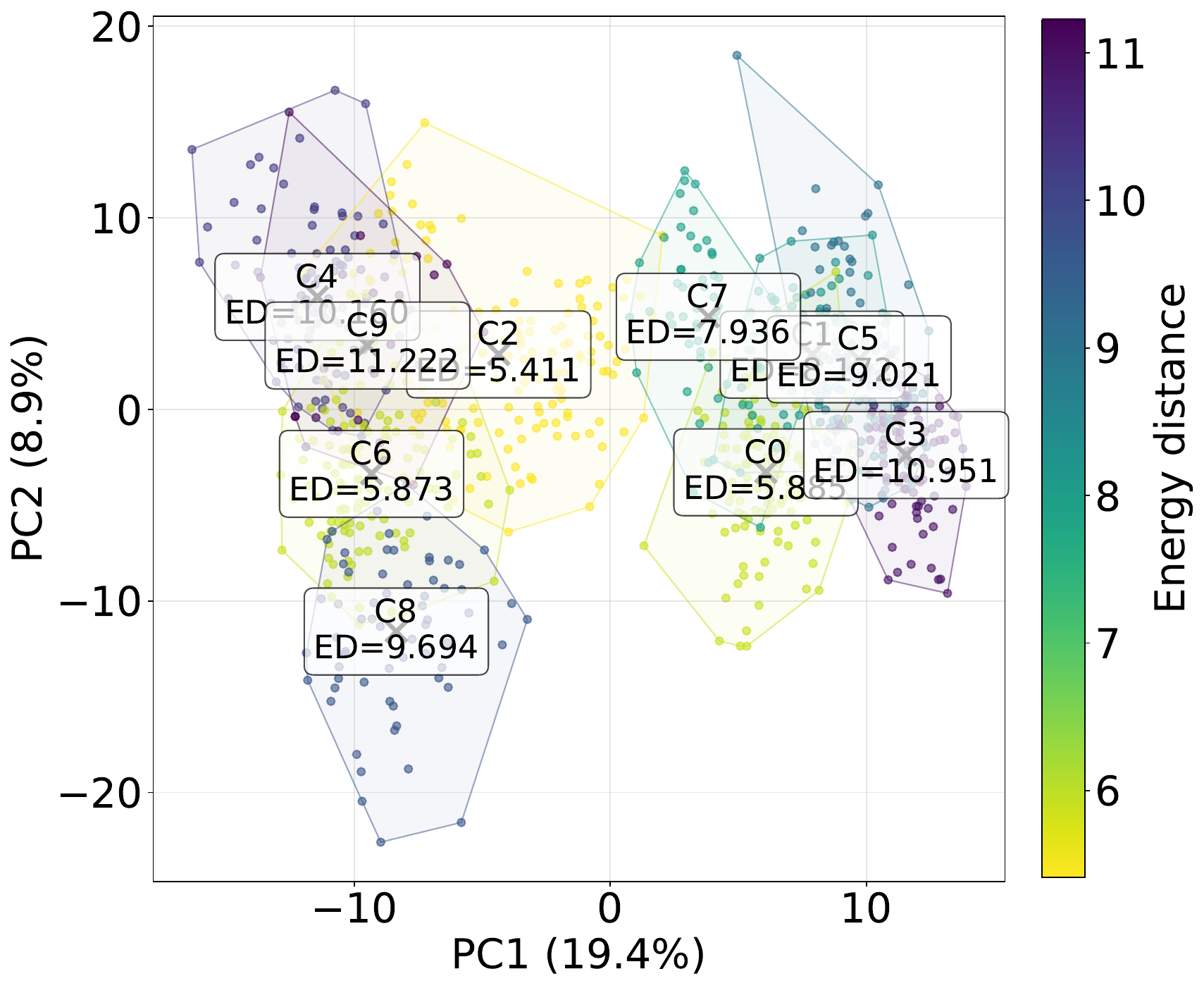}
    \end{subfigure}

    \vspace{0.5em}

    \begin{subfigure}[t]{0.48\linewidth}
        \centering
        \includegraphics[width=\linewidth]{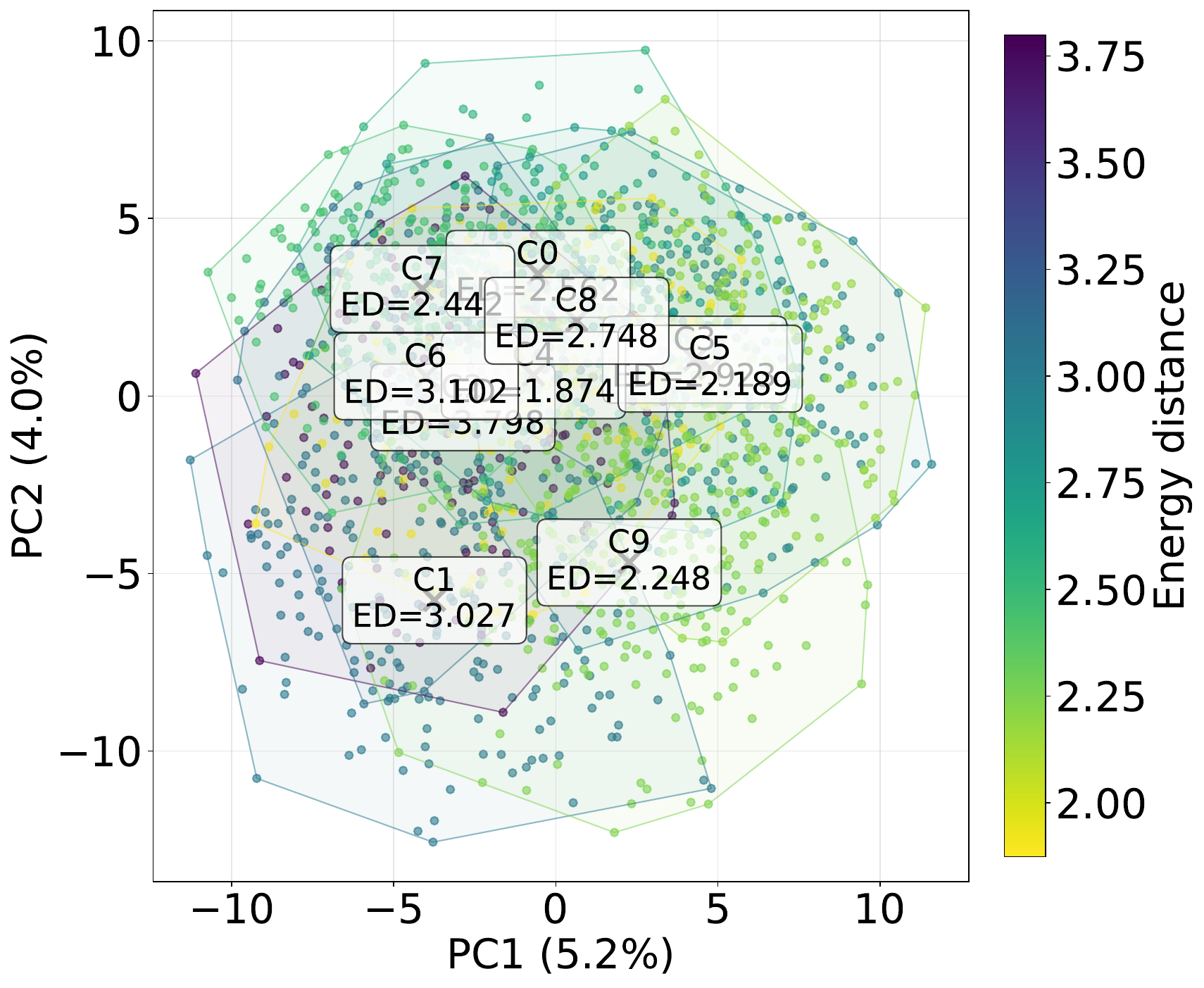}
    \end{subfigure}
    \hfill
    \begin{subfigure}[t]{0.48\linewidth}
        \centering
        \includegraphics[width=\linewidth]{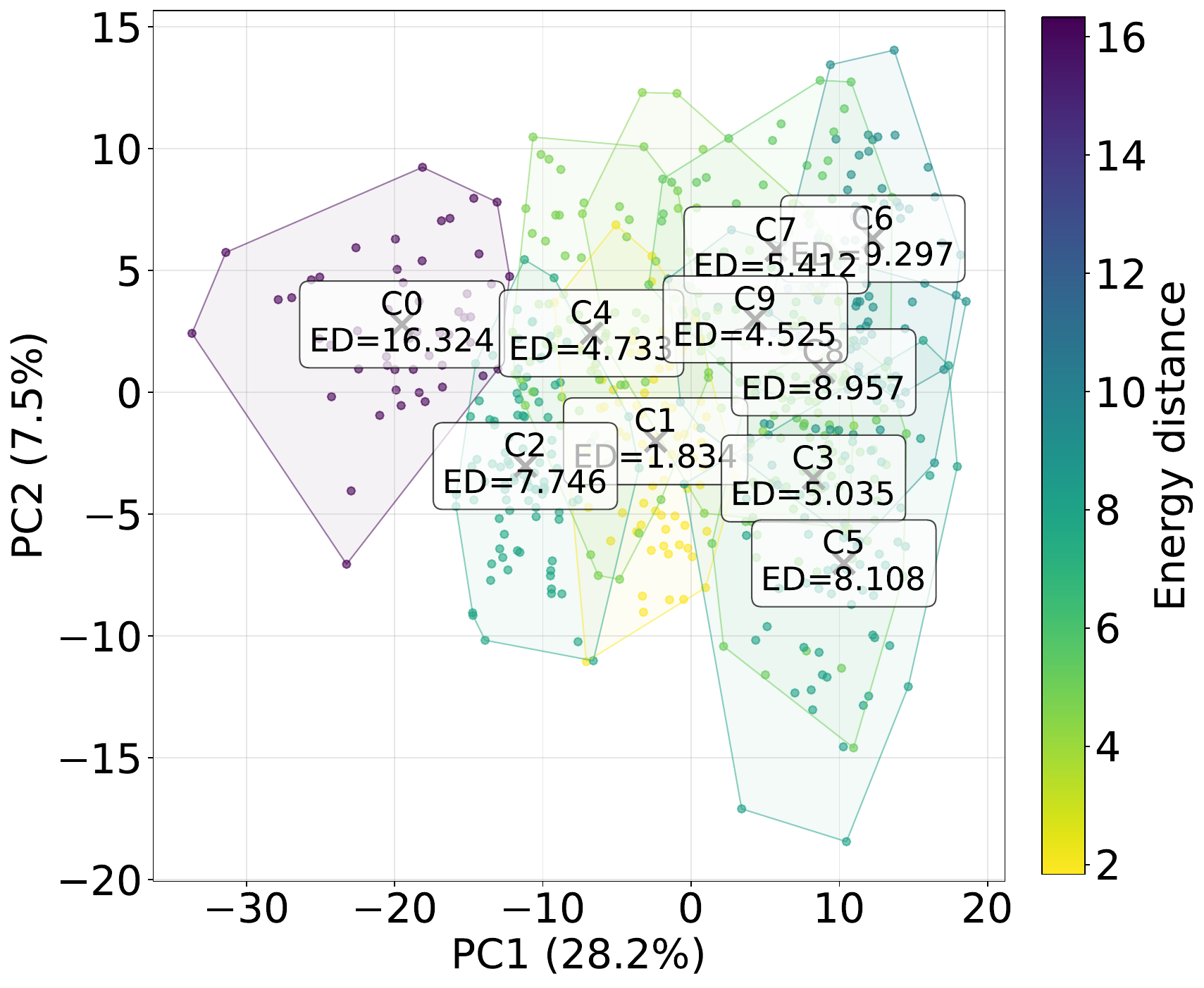}
    \end{subfigure}

    \caption{\small
    \textbf{Preliminary structural comparison between free-form text and serialized tabular inputs.}
    Left: free-form text reference datasets. Right: serialized tabular datasets.
    PCA visualizations in an external embedding space show that serialized tabular inputs exhibit more structured local neighborhoods, reflecting repeated schema slots and similar attribute-value patterns.
    }
    \label{fig:kmean}
    \vspace{-5mm}
\end{figure}

Machine unlearning aims to remove the influence of designated training
data from a model without retraining from scratch~\cite{
cao2015towards,thudi2022unrolling,jang2023knowledge,yao2023large,
liu2025rethinking}. 
Recent LLM unlearning work has largely focused on free-form text or
knowledge-oriented settings, where methods modify the forgetting
objective, add regularization, or design specialized parameter
updates~\cite{jang2023knowledge,yao2023large,
li2024wmdpbenchmarkmeasuringreducing,zhang2024negative,
liu2025rethinking}. 
However, unlearning for tabular data remains underexplored. 
This gap is important because tabular prediction is a common format for
real-world decision pipelines, and recent work increasingly adapts
tabular records to language models through record serialization,
schema-aware prompting, and table representation
learning~\cite{hegselmann2023tabllmfewshotclassificationtabular,
carballo2025tabtextlanguagebasedrepresentationstabular,
herzig2020tapas,liu2021tapex,wu2025tabular}. 
In this paper, we focus on unlearning serialized tabular records or
attributes from LLMs while preserving useful behavior on retained data.

The key challenge that distinguishes tabular unlearning is
\emph{overlap}. 
When tabular records are serialized, every row is expressed under the same schema: repeated column names, similar attribute order, shared value ranges, and a common output space. Thus, two different records may remain locally close to the model even when they belong to different splits. We call this \emph{schema-induced overlap}. 
Figure ~\ref{fig:kmean} provides a qualitative diagnostic: in an external embedding space, the serialized tabular datasets considered here exhibit more regular local neighborhoods than the free-form text references, plausibly reflecting repeated column-name/value slots and similar attribute-value patterns. This motivates the possibility that a forget row may lie near many retained rows under the shared schema.

This property directly affects retain-preserving unlearning. 
A standard formulation partitions training data into a forget set,
whose influence should be removed, and a retain set, whose behavior
should be preserved~\cite{li2024wmdpbenchmarkmeasuringreducing,
zhang2024negative}. 
Such objectives are easier to optimize when forget and retain examples
are well separated. 
In serialized tabular data, however, a forgotten row may be close to
retained rows under the shared schema. 
Preserving all such nearby retained rows can restrict the model changes
needed to remove the influence of the forget set. 
Prior work shows that data correlation, representation overlap, and
conflicting update directions can degrade the forget--utility
trade-off~\cite{zhao2024makes,
golatkar2020eternalsunshinespotlessnet,
huang2024unifiedgradientbasedmachineunlearning,
reisizadeh2025blurbileveloptimizationapproach}. 
For tabular LLMs, we argue that this interference can arise naturally
from the shared schema itself.
This raises the central question of this paper:

\begin{mdframed}[userdefinedwidth=.99\linewidth,align=center,skipabove=3pt,skipbelow=3pt,innerleftmargin=6pt,innerbottommargin=6pt,innertopmargin=6pt,roundcorner=3pt,backgroundcolor=cyan!5,linecolor=gray]  
\noindent \textbf{RQ1.}
\textit{How can LLMs unlearn tabular records or attributes while
preserving retain behavior when fixed schemas induce overlap between
the forget and retain sets?}
\end{mdframed}

Motivated by this question, we propose
\texttt{Conflict-Aware Unlearning} (\methodname), a schema-aware
retain-construction method. 
Instead of preserving every retain row with equal strength,
\methodname\ identifies retain rows that lie close to the forget
distribution in schema-aligned attribute space and relaxes their
preservation constraints during optimization. 
Rows far from the forget distribution remain strong preservation
anchors. 
Importantly, \methodname\ does not change the forgetting objective or
the optimizer; it changes how retain constraints are constructed.

We evaluate \methodname\ under both sample-level and feature-level
tabular unlearning. 
Sample-level unlearning removes entire records, while feature-level
unlearning suppresses selected attribute-level evidence spans. 
Across overlap regimes, model sizes, and tabular prediction tasks,
\methodname\ more closely matches a retraining oracle than general LLM
unlearning baselines while preserving predictive utility and
retain-region behavior. 
These results show that reliable tabular LLM unlearning requires
explicitly accounting for schema-induced forget--retain overlap.
Our contributions are as follows:
\begin{itemize}
    \item We study LLM unlearning for serialized tabular data and identify \emph{schema-induced forget--retain overlap} as a key challenge in this underexplored setting, where fixed schemas and repeated column-name/value slots can make forget examples locally close to retained examples.
    \vspace{-2mm}
    \item We propose \methodname, a schema-aware retain-construction method that relaxes high-conflict retain constraints before optimization.
    \vspace{-2mm}
    \item We evaluate CAU under sample-level and feature-level tabular unlearning, showing that conflict-aware retain construction improves oracle agreement most clearly in high-overlap regimes and identifying lower-overlap cases where simpler retain-preserving methods are sufficient.
\end{itemize}

\section{Related Work}
\label{sec:related_work}
\noindent\textbf{Machine Unlearning.} Machine unlearning studies how to remove the influence of a designated forget set from a trained model while preserving performance on retained data, serving as an efficient alternative to full retraining under deletion requests~\cite{jang2023knowledge,yao2023large,liu2025rethinking}.
Prior work spans training-time designs such as data sharding or slicing~\cite{bourtoule2020machineunlearning,ginart2019makingaiforgetyou} and post-hoc approaches that approximate retraining via fine-tuning, distillation, or regularization-based updates~\cite{golatkar2020eternalsunshinespotlessnet,guo2023certifieddataremovalmachine,ding2025understandingfinetuningapproximateunlearning}.
Recent studies further extend unlearning to large models and LLMs, emphasizing empirical evaluation of forgetting effectiveness and retain utility~\cite{eldan2023whosharrypotterapproximate,peng2025forgetknowrememberuse}.
Membership inference has likewise exposed privacy leakage in other generative-model families, including diffusion models~\cite{duan2023diffusion}, supporting its use as a general privacy-auditing mechanism.

\noindent\textbf{LLMs for Tabular and Structured Data.}
Recent work has explored adapting large language models to tabular and structured data through serialization, schema-aware prompting, and hybrid encoders (e.g., TabText)~\cite{carballo2025tabtextlanguagebasedrepresentationstabular,liu2022tapextablepretraininglearning,wu2025tabulardataunderstandingllms,Herzig_2020}. These approaches demonstrate that LLMs can effectively process tabular inputs but also show that fixed schemas and repeated attribute tokens induce strong representation regularities~\cite{deng2020turltableunderstandingrepresentation}. Our analysis builds on this line of work and shows that such schema-driven regularity has direct implications for unlearning, leading to stronger forget–retain coupling than typically observed in free-form text settings.
Complementary evaluations probe relevance-aware predictive uncertainty in free-form LLM outputs and strategic reasoning through game-theoretic tasks~\cite{duan2024shifting,duan2024gtbench}; our study instead asks how unlearning behaves for schema-serialized prediction.

\noindent\textbf{Unlearning under Data Correlation.}
A growing number of work shows that unlearning performance degrades substantially when forget and retain data are statistically or representationally entangled~\cite{zhao2024makes}.
In such settings, unlearning updates may inadvertently propagate to nearby retain samples, leading to unfavorable forget-utility trade-offs~\cite{zhao2024makes,golatkar2020forgettingoutsideboxscrubbing,ding2025understandingfinetuningapproximateunlearning}.
These observations motivate structure-aware unlearning strategies that explicitly account for data overlap and correlation during targeted removal~\cite{neel2020descenttodeletegradientbasedmethodsmachine,guo2023certifieddataremovalmachine}.

\vspace{-2mm}
\section{Motivating Analysis}
\vspace{-2mm}
\label{sec:prelim}
Serialized tabular inputs share a fixed prompt schema across all records: every row expresses its content through the same column-name/value slots, and records with similar values in high-signal columns may become close in representation space and may exhibit similar predictive behavior, a pattern we call \emph{schema-induced overlap}.
However, structural proximity alone does not fully explain why unlearning becomes difficult. A nearby retain row becomes a conflicting constraint only when it  preserves behavior that depends on the same attribute-level evidence  that the forgetting update must modify. We therefore use this section to examine the behavioral side of the problem: whether serialized tabular predictions rely on repeated attribute-level evidence, and how this can turn local overlap into forget--retain interference.

\subsection{Schema-Aligned Predictive Evidence}
\label{sec:prelim_feature}

The PCA diagnostic in Fig.~\ref{fig:kmean} suggests that serialized tabular inputs form more structured local neighborhoods than free-form text references. We next ask whether this structural proximity is behaviorally relevant: do nearby tabular records tend to rely on repeated schema-aligned evidence?

To measure token-level evidence, we compute gradient-based saliency with respect to the input embeddings. Let $E(x)\in\mathbb{R}^{T\times d}$ denote the input embedding sequence, and let $\ell(x,y)$ be the negative log-likelihood loss of the target label token. For each token position $t$, we define
\begin{equation}
s_t(x)=\left\lVert \nabla_{E_t(x)} \ell(x,y) \right\rVert_2,
\qquad t=1,\dots,T .
\end{equation}

\begin{figure}[t]
    \centering

    \begin{subfigure}[t]{0.48\textwidth}
        \centering
        \includegraphics[width=\linewidth]{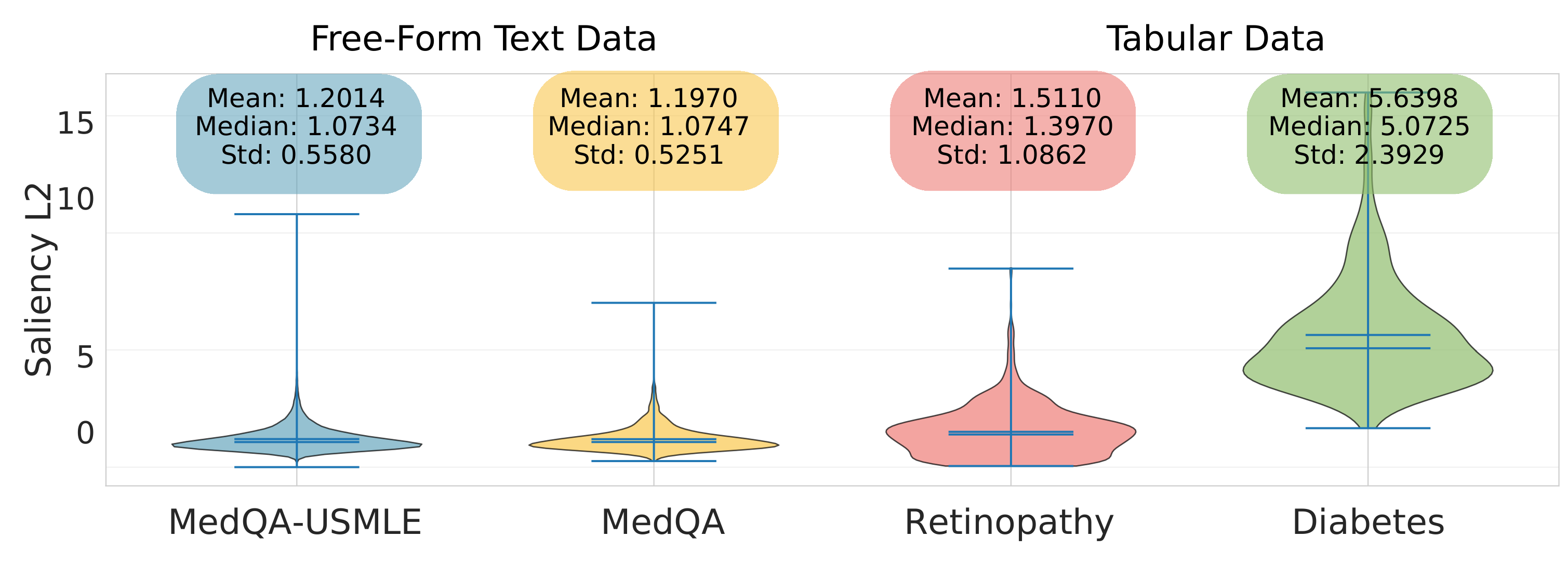}
        \caption{\small Token-level saliency.}
        \label{fig:saliency_violin}
    \end{subfigure}
    \hfill
    \begin{subfigure}[t]{0.48\textwidth}
        \centering
        \includegraphics[width=\linewidth]{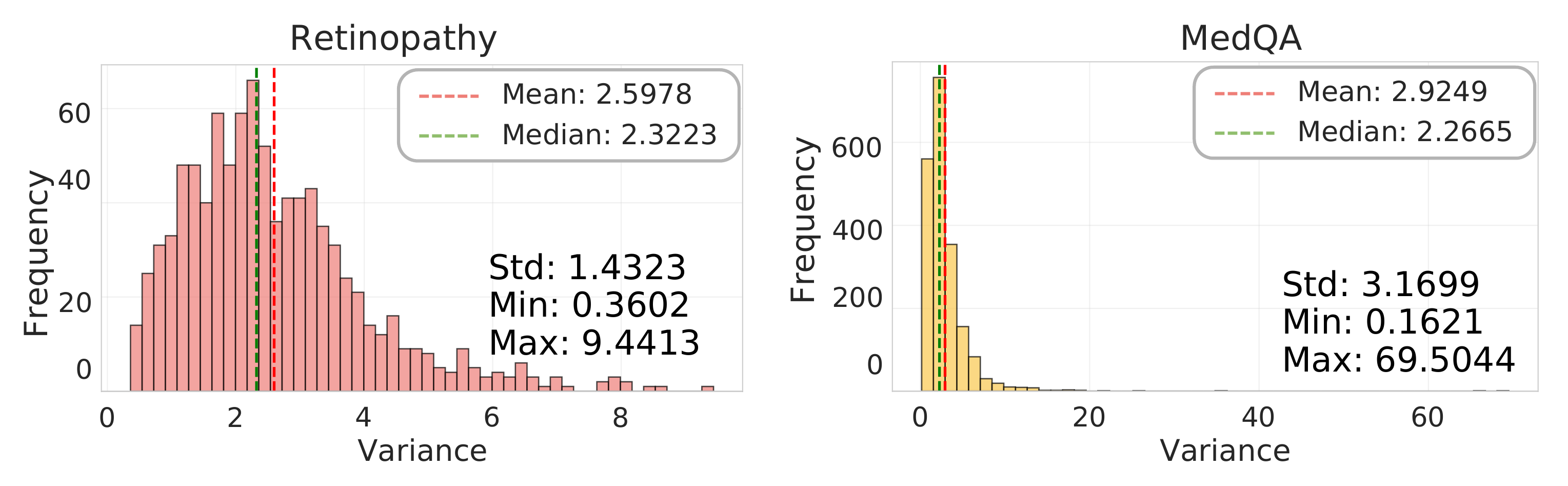}
        \caption{\small Perturbation sensitivity.}
        \label{fig:saliency_violin2}
    \end{subfigure}

    \caption{\small
    \textbf{Schema-aligned evidence and perturbation sensitivity across modalities.}
    Token-level saliency shows that serialized tabular predictions place strong evidence on repeated schema-aligned tokens, such as column names and corresponding values. 
    Perturbation diagnostics measure how sensitive the target-label logit is to small input-form changes.
    Together, these analyses suggest that tabular predictions can depend on recurring attribute-level evidence shared across records.
    }
    \label{fig:structured_feature}
    \vspace{-4mm}
\end{figure}

As shown in Fig.~\ref{fig:structured_feature}(a), tabular inputs show strong saliency on schema-aligned tokens, including column names and their associated values. This suggests that model predictions on serialized tabular prompts are often mediated by repeated attribute-level evidence rather than by free-form semantic variation alone. Since the same columns recur across records, two nearby rows may depend on overlapping evidence even when they correspond to different samples.

We further examine sensitivity to small input-form perturbations. Let $g(x)$ denote the model logit of the correct label token at the final position. For each sample $x$, we generate $K$ input variants 
$\{x^{(k)}\}_{k=1}^{K}$ and compute
\begin{equation}
\begin{aligned}
\mathrm{Var}(x)
&=
\frac{1}{K}\sum_{k=1}^{K}
\left(g(x^{(k)})-\mu(x)\right)^2,\\
\mu(x)
&=
\frac{1}{K}\sum_{k=1}^{K} g(x^{(k)}).
\end{aligned}
\end{equation}

This perturbation diagnostic measures whether small changes in input form can substantially alter the model's prediction. In serialized tabular prompts, such sensitivity is especially relevant because small perturbations often affect repeated schema slots or attribute descriptions. Together with the saliency analysis, this indicates that tabular predictions can be strongly tied to recurring schema-level evidence.

\subsection{From Schema-Induced Overlap to Unlearning Conflict}
\label{sec:failure_mode}

The preceding diagnostics identify two properties of serialized tabular prediction. First, records can form local neighborhoods under a shared schema, so a forget sample may have nearby retain samples. Second, predictions may rely on repeated schema-aligned evidence, so nearby records can depend on the same columns or attribute-value patterns. These two properties jointly create the failure mode studied in this paper.

Most retain-preserving unlearning methods enforce preservation on all retain samples during forgetting. This is reasonable when forget and retain samples are sufficiently separable. In high-overlap tabular settings, however, a nearby retain sample may ask the model to preserve precisely the schema-aligned behavior that the forget update needs to change. Consequently, the forgetting loss and the retain-preservation loss can impose opposing constraints on the same representations. This does not mean that nearby retain samples are unimportant for evaluation; rather, it means that enforcing all of them equally during optimization may make the unlearning objective ill-conditioned. This leads to the following hypothesis: \textit{In schema-structured tabular settings, standard retain-preserving unlearning can become unstable when forget samples have nearby retain neighbors that rely on overlapping attribute-level evidence.}
% \begin{mdframed}[userdefinedwidth=.99\linewidth,align=center,skipabove=3pt,skipbelow=3pt,innerleftmargin=6pt,innerbottommargin=6pt,innertopmargin=6pt,roundcorner=3pt,backgroundcolor=blue!5,linecolor=gray]
% \noindent \textbf{Hypothesis.} 
% \textit{In schema-structured tabular settings, standard retain-preserving unlearning can become unstable when forget samples have nearby retain neighbors that rely on overlapping attribute-level evidence.}
% \end{mdframed}

\section{Method: \texttt{Conflict-Aware Unlearning} }
\label{sec:method}
The analysis in Sec.~\ref{sec:prelim} suggests that forget--retain 
overlap can turn retain preservation into an optimization constraint 
that opposes forgetting, motivating the following question:

\begin{mdframed}[userdefinedwidth=.99\linewidth,align=center,
skipabove=3pt,skipbelow=3pt,innerleftmargin=6pt,
innerbottommargin=6pt,innertopmargin=6pt,roundcorner=3pt,
backgroundcolor=cyan!5,linecolor=gray]  
\noindent \textbf{RQ 2:} \emph{How can we design an unlearning method that reduces 
interference between forget and retain samples while maintaining effective 
forgetting and predictive performance?}
\end{mdframed}

To answer RQ2, \methodname keeps the forgetting target unchanged but modifies 
retain-constraint strength according to proximity to the forget distribution, 
downweighting or filtering high-conflict retain rows during optimization.

\subsection{Schema-Aware Distance to the Forget Distribution}
\label{sec:metric}

To identify retain rows that may conflict with forgetting, we first 
measure how close each row is to the forget-set distribution in 
tabular attribute space. 
Let $\mathcal{A}$ denote the fixed tabular schema. 
Each serialized record $x$ is parsed into attribute--value pairs 
$\{(a,v_a(x)) : a\in\mathcal{A}(x)\}$, where 
$\mathcal{A}(x)\subseteq\mathcal{A}$ is the set of attributes present 
in $x$. 
Since tabular datasets may contain both numerical and categorical 
columns, we associate each attribute $a$ with a type-specific encoder
\[
\phi_a(v_a(x))\in\mathbb{R}^{d_a}.
\]
For numerical attributes, $\phi_a$ returns the scalar value. 
For categorical attributes, $\phi_a$ returns a one-hot representation 
or an equivalent fixed categorical encoding. 
Thus, all attributes are represented in a schema-aligned value space, 
rather than in raw text space.

For each attribute $a$, we estimate the forget-set mean and standard 
deviation in this encoded space:
\begin{equation}
\begin{aligned}
\mu_a^{(f)}
&=
\frac{1}{|\mathcal{D}_f(a)|}
\sum_{x\in\mathcal{D}_f(a)}
\phi_a(v_a(x)),
\\[6pt]
\sigma_a^{(f)}
&=
\sqrt{
\frac{1}{|\mathcal{D}_f(a)|}
\sum_{x\in\mathcal{D}_f(a)}
\left(
\phi_a(v_a(x))-\mu_a^{(f)}
\right)^2
},
\end{aligned}
\label{eq:forget_stats}
\end{equation}
where 
$\mathcal{D}_f(a)=\{x\in\mathcal{D}_f: a\in\mathcal{A}(x)\}$.
The square and square root are applied elementwise when 
$d_a>1$.

We then define an attribute-level normalized distance:
\begin{equation}
\delta_a(x;\mathcal{D}_f)
=
\frac{1}{d_a}
\left\|
\frac{
\phi_a(v_a(x))-\mu_a^{(f)}
}{
\max\{\sigma_a^{(f)},\epsilon\}
}
\right\|_2^2,
\label{eq:attr_distance}
\end{equation}
where $\epsilon>0$ avoids degenerate normalization and the division is 
elementwise. 
Let
\[
\mathcal{S}(x)
=
\{a\in\mathcal{A}(x): |\mathcal{D}_f(a)|>0\}
\]
be the attributes of $x$ whose forget-set statistics are defined. 
The \texttt{Forget-Normalized Distance} (FND) of $x$ is
\begin{equation}
\mathrm{FND}(x;\mathcal{D}_f)
=
\sqrt{
\frac{1}{|\mathcal{S}(x)|}
\sum_{a\in\mathcal{S}(x)}
\delta_a(x;\mathcal{D}_f)
}.
\label{eq:fnd}
\end{equation}

For all-numerical datasets, Eq.~\eqref{eq:fnd} reduces to the 
root-mean-square of forget-normalized $z$-scores across columns, 
equivalently a diagonal Mahalanobis distance under the forget-set 
empirical statistics. 
A smaller $\mathrm{FND}(x;\mathcal{D}_f)$ means that $x$ is closer to 
the forget distribution under the shared tabular schema, and is 
therefore more likely to act as a high-conflict retain constraint.

\subsection{Conflict-Aware Retain Construction}
\label{sec:carf}

Given FND scores, \methodname\ constructs retain constraints according 
to their proximity to the forget distribution. 
We first compute the forget-set distance distribution
\[
\mathcal{V}_f
=
\{\mathrm{FND}(x_f;\mathcal{D}_f):
x_f\in\mathcal{D}_f,\ \mathcal{S}(x_f)\neq\emptyset\}.
\]
For a percentile $p\in(0,100)$, we define the adaptive cutoff
\begin{equation}
\tau_p
=
\operatorname{Percentile}(\mathcal{V}_f,p).
\label{eq:tau}
\end{equation}

Retain rows whose distance to the forget distribution is below this 
cutoff are treated as high-conflict constraints:
\begin{align}
\widetilde{\mathcal{D}}_r
&=
\{x_r\in\mathcal{D}_r:
\mathcal{S}(x_r)\neq\emptyset\},
\label{eq:valid_retain}
\\
\mathcal{D}_r^{\mathrm{drop}}
&=
\{x_r\in\widetilde{\mathcal{D}}_r:
\mathrm{FND}(x_r;\mathcal{D}_f)\le \tau_p\},
\label{eq:drop}
\\
\mathcal{D}_r^{\mathrm{safe}}
&=
\mathcal{D}_r\setminus\mathcal{D}_r^{\mathrm{drop}}.
\label{eq:safe}
\end{align}

More generally, we assign each retain row a constraint weight
\begin{equation}
w_{\lambda,p}(x_r)
=
\begin{cases}
\lambda, & x_r\in\mathcal{D}_r^{\mathrm{drop}},\\
1, & x_r\in\mathcal{D}_r^{\mathrm{safe}},
\end{cases}
\label{eq:retain_weight}
\end{equation}
where $\lambda\in[0,1]$ controls how strongly high-conflict retain rows 
are preserved. 
When $\lambda=0$, \methodname\ reduces to hard filtering and removes 
$\mathcal{D}_r^{\mathrm{drop}}$ from the retain-preservation loss. 
When $0<\lambda<1$, \methodname\ softly downweights high-conflict 
retain rows. 
When $\lambda=1$, the method recovers the standard retain-all objective.

To control for the effect of reducing the number of retain constraints, we also use a random retain-selection control in our experiments:
\begin{equation}
\mathcal{D}_r^{\mathrm{rand}}
\sim
\operatorname{Unif}
\left(
\left\{
\mathcal{S}\subseteq\widetilde{\mathcal{D}}_r:
|\mathcal{S}|=|\mathcal{D}_r^{\mathrm{safe}}|
\right\}
\right).
\label{eq:random_control}
\end{equation}
This control retains the same number of rows as hard \methodname but selects them without using forget-set proximity.

\subsection{Instantiating CAU in Retain-Preserving Unlearning}
\label{sec:rmu_integration}

\methodname\ can be plugged into any unlearning objective with a 
retain-preservation term. 
In this work, we instantiate it with a representation-level objective. 
Let $f_{\theta}$ be the model being unlearned and $f_{\theta_0}$ be the 
frozen reference model before unlearning. 
For input $x$, let $\mathcal{T}(x)$ denote its token positions and 
$M_{\theta}^{(\ell)}(t)$ the hidden state at layer $\ell$.

To support different deletion granularities, we define 
$\mathcal{T}_f(x)$ as the token positions targeted for forgetting. 
For sample-level unlearning, $\mathcal{T}_f(x)=\mathcal{T}(x)$. 
For feature-level unlearning, $\mathcal{T}_f(x)$ contains only the 
token spans of selected attribute names and values. 
Retain preservation is computed on the full retained row.

The forget and retain losses are
\begin{align}
\ell_f(x;\theta)
&=
\frac{1}{|\mathcal{T}_f(x)|}
\sum_{t\in\mathcal{T}_f(x)}
\left\|
M_{\theta}^{(\ell)}(t)-c\mathbf{u}
\right\|_2^2,
\label{eq:forget_loss}
\\
\ell_r(x;\theta)
&=
\frac{1}{|\mathcal{T}(x)|}
\sum_{t\in\mathcal{T}(x)}
\left\|
M_{\theta}^{(\ell)}(t)-
M_{\theta_0}^{(\ell)}(t)
\right\|_2^2,
\label{eq:retain_loss}
\end{align}
where $\mathbf{u}$ is a fixed random unit vector and $c>0$ controls the 
forgetting strength.

CAU modifies only the retain side of the objective. 
Instead of preserving all retain samples uniformly, it computes the 
retain loss on the low-conflict subset 
$\mathcal{D}_r^{\mathrm{safe}}$:
\begin{equation}
\begin{aligned}
\min_{\theta}\quad 
\mathcal{L}_{\mathrm{CAU}}(\theta)
&=
\mathbb{E}_{x_f\sim\mathcal{D}_f}
[\ell_f(x_f;\theta)] \\
&\quad+
\alpha\,
\mathbb{E}_{x_r\sim\mathcal{D}_r^{\mathrm{safe}}}
[\ell_r(x_r;\theta)] .
\end{aligned}
\label{eq:cau_objective}
\end{equation}
Here $\alpha>0$ controls the strength of retain preservation. 
The forgetting loss and its target set are unchanged; CAU only replaces 
the retain constraints with a lower-conflict retain subset.

\paragraph{Soft weighting.}
A soft variant replaces hard filtering with weights 
$w_{\lambda,p}(x)=\lambda$ for 
$x\in\mathcal{D}_r^{\mathrm{drop}}$ and 
$w_{\lambda,p}(x)=1$ for 
$x\in\mathcal{D}_r^{\mathrm{safe}}$, where 
$\lambda\in[0,1]$. 
We evaluate this variant in the ablation study.

\paragraph{Optimization.}
We compute FND scores and construct 
$\mathcal{D}_r^{\mathrm{safe}}$ once before unlearning. 
During training, minibatches draw forget samples from $\mathcal{D}_f$ 
and retain samples from $\mathcal{D}_r^{\mathrm{safe}}$. 
Thus, \methodname\ adds only a one-time schema-level scoring step and 
does not introduce additional backward passes.

\begin{table*}[t]
\centering
\scriptsize

% ================= LEFT TABLE =================
\begin{minipage}{0.54\textwidth}
\raggedleft
\setlength{\tabcolsep}{2pt}
\renewcommand{\arraystretch}{1.05}

\begin{tabular}{lcccccc}
\toprule
& \multicolumn{6}{c}{\textbf{Qwen2.5 3B Instruct}} \\
\cmidrule(lr){2-7}
& \multicolumn{3}{c}{High-overlap} & \multicolumn{3}{c}{Low-overlap} \\
\cmidrule(lr){2-4} \cmidrule(lr){5-7}
Method
& MIA AUC & $|\Delta|$ & Test Acc
& MIA AUC & $|\Delta|$ & Test Acc \\
\midrule
Raw Model
& 0.3507 & 0.1627 & 0.6147
& 0.5851 & 0.0700 & 0.6147 \\
Tuned Model
& 0.2433 & 0.2701 & 0.6494
& 0.5781 & 0.0630 & 0.6494 \\
\rowcolor{green!15}
Retrained (Oracle)
& 0.5134 & 0 & 0.6667
& 0.5151 & 0 & 0.6926 \\
\midrule
GradDiff
& 0.2304 & 0.2830 & 0.6450
& 0.5978 & 0.0827 & 0.6450 \\
NPO
& 0.2432 & 0.2702 & 0.6061
& 0.5937 & 0.0786 & 0.6364 \\
SimNPO
& 0.3865 & 0.1269 & 0.6450
& 0.5981 & 0.0830 & 0.6450 \\
RMU
& 0.2799 & 0.2335 & 0.6407
& 0.5809 & 0.0658 & 0.6536 \\
\rowcolor{red!15}
\textbf{Ours}
& \textbf{0.4546} & \textbf{0.0588} & 0.9957
& \textbf{0.4986} & \textbf{0.0165} & 0.6450 \\
\midrule\midrule
& \multicolumn{6}{c}{\textbf{Qwen2.5 7B Instruct}} \\
\cmidrule(lr){2-7}
& \multicolumn{3}{c}{High-overlap} & \multicolumn{3}{c}{Low-overlap} \\
\cmidrule(lr){2-4} \cmidrule(lr){5-7}
Method
& MIA AUC & $|\Delta|$ & Test Acc
& MIA AUC & $|\Delta|$ & Test Acc \\
\midrule
Raw Model
& 0.9747 & 0.5417 & 0.8701
& 0.8929 & 0.4747 & 0.8701 \\
Tuned Model
& 0.9763 & 0.5433 & 0.9177
& 0.8989 & 0.4807 & 0.9264 \\
\rowcolor{green!15}
Retrained (Oracle)
& 0.4330 & 0 & 0.7792
& 0.4182 & 0 & 0.7792 \\
\midrule
GradDiff
& 0.9782 & 0.5452 & 0.9134
& 0.9023 & 0.4841 & 0.9264 \\
NPO
& 0.9675 & 0.5345 & 0.9264
& 0.9046 & 0.4864 & 0.9221 \\
SimNPO
& 0.9699 & 0.5369 & 0.9177
& 0.8919 & 0.4737 & 0.9177 \\
RMU
& 0.3142 & 0.1188 & 0.6061
& 0.1514 & 0.2668 & 0.9957 \\
\rowcolor{red!15}
\textbf{Ours}
& \textbf{0.5390} & \textbf{0.1060} & 0.8139
& \textbf{0.5473} & \textbf{0.1291} & 0.8355 \\
\bottomrule
\end{tabular}

\vspace{3pt}
% \caption*{\textbf{Retinopathy}}
\end{minipage}
\hspace{0.005\textwidth}
% ================= RIGHT TABLE =================
\begin{minipage}{0.44\textwidth}
\raggedright
\setlength{\tabcolsep}{2pt}
\renewcommand{\arraystretch}{1.05}

\begin{tabular}{cccccc}
\toprule
\multicolumn{6}{c}{\textbf{Qwen2.5 3B Instruct}} \\
\cmidrule(lr){1-6}
\multicolumn{3}{c}{High-overlap} & \multicolumn{3}{c}{Low-overlap} \\
\cmidrule(lr){1-3} \cmidrule(lr){4-6}
MIA AUC & $|\Delta|$ & Test Acc
& MIA AUC & $|\Delta|$ & Test Acc \\
\midrule
0.4667 & 0.0498 & 0.6753 & 0.5487 & 0.0542 & 0.6753 \\
0.5096 & 0.0069 & 0.7922 & 0.2891 & 0.2054 & 0.7922 \\
\rowcolor{green!15}
0.5165 & 0 & 0.7403 & 0.4945 & 0 & 0.7143 \\
\midrule
0.0093 & 0.5072 & 0.3571 & 0.0287 & 0.4658 & — \\
0.2508 & 0.2657 & 0.7597 & 0.0199 & 0.4746 & — \\
0.0038 & 0.5127 & — & 0 & 0.4945 & — \\
0.4676 & 0.0489 & 0.6429 & \textbf{0.4812} & \textbf{0.0133} & 0.3571 \\
\rowcolor{red!15}
\textbf{0.5332} & \textbf{0.0167} & 0.7922 & 0.4712 & 0.0233 & 0.7597 \\
\midrule\midrule
\multicolumn{6}{c}{\textbf{Qwen2.5 7B Instruct}} \\
\cmidrule(lr){1-6}
\multicolumn{3}{c}{High-overlap} & \multicolumn{3}{c}{Low-overlap} \\
\cmidrule(lr){1-3} \cmidrule(lr){4-6}
MIA AUC & $|\Delta|$ & Test Acc
& MIA AUC & $|\Delta|$ & Test Acc \\
\midrule
0.5954 & 0.1078 & 0.6494 & 0.5667 & 0.0905 & 0.6494 \\
0.6194 & 0.1318 & 0.7338 & 0.5661 & 0.0899 & 0.7338 \\
\rowcolor{green!15}
0.4876 & 0 & 0.6429 & 0.4762 & 0 & 0.6429 \\
\midrule
0.5923 & 0.1047 & — & 0.5315 & 0.0553 & 0.3571 \\
0.5927 & 0.1051 & 0.3571 & 0.5294 & 0.0532 & — \\
0.5934 & 0.1058 & 0.6429 & 0.5301 & 0.0539 & — \\
0.5631 & 0.0755 & 0.6818 & 0.5314 & 0.0552 & 0.6948 \\
\rowcolor{red!15}
\textbf{0.5305} & \textbf{0.0429} & 0.7597 & \textbf{0.4978} & \textbf{0.0216} & 0.7597 \\
\bottomrule
\end{tabular}

\vspace{3pt}
% \caption*{\textbf{Diabetes}}
\end{minipage}

\caption{\small 
\textbf{Feature-level unlearning on Retinopathy (left) and Diabetes (right).} Results are shown under \textsc{high-overlap} and \textsc{low-overlap} regimes. \textsc{Retrained (Oracle)} is trained only on $\mathcal{D}_r$, and $|\Delta|$ denotes deviation from the oracle MIA AUC. ``---'' indicates invalid prediction outputs.
}
\label{tab:combined}
\vspace{-3mm}
\end{table*}

\section{Experiments}
\label{sec:exp}
\begin{figure*}[t]
    \centering

    \begin{subfigure}[t]{0.48\textwidth}
        \centering
        \includegraphics[width=\linewidth]{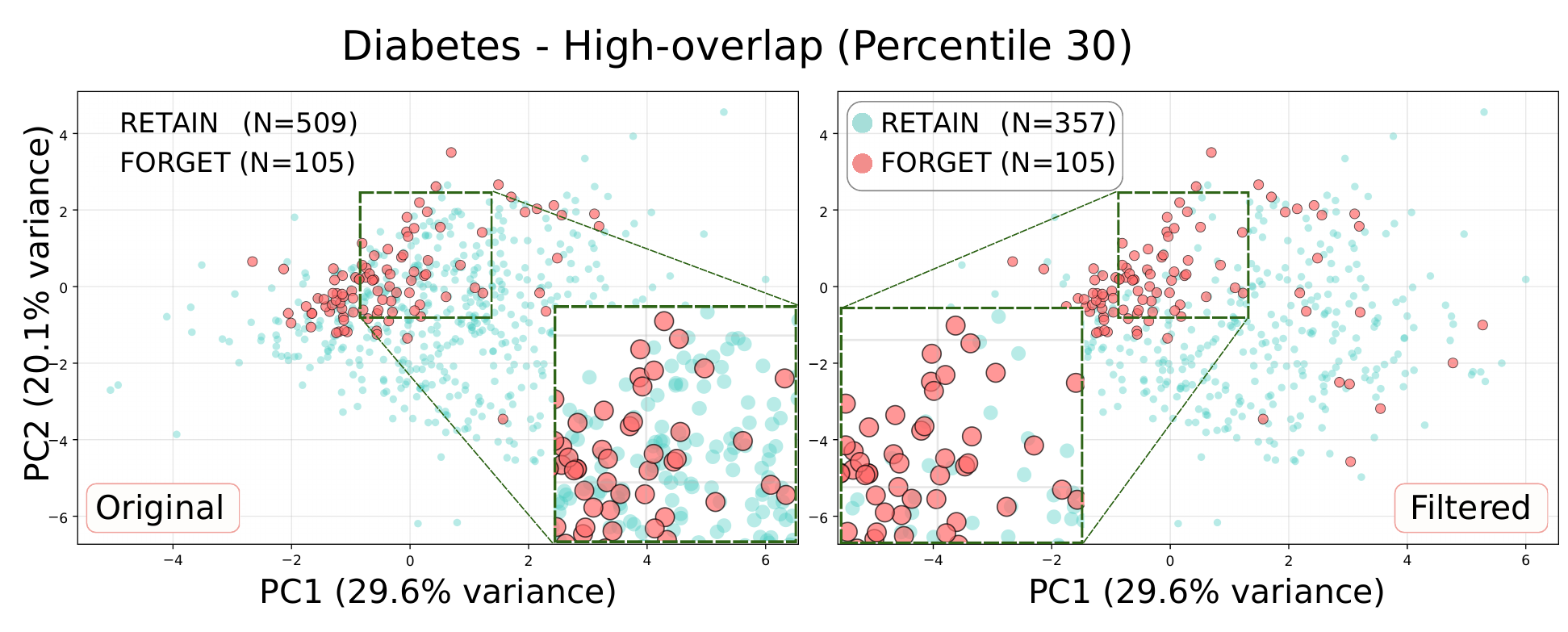}
        \caption{\textsc{High-overlap}}
        \label{fig:diabetes_change_high}
    \end{subfigure}
    \hfill
    \begin{subfigure}[t]{0.48\textwidth}
        \centering
        \includegraphics[width=\linewidth]{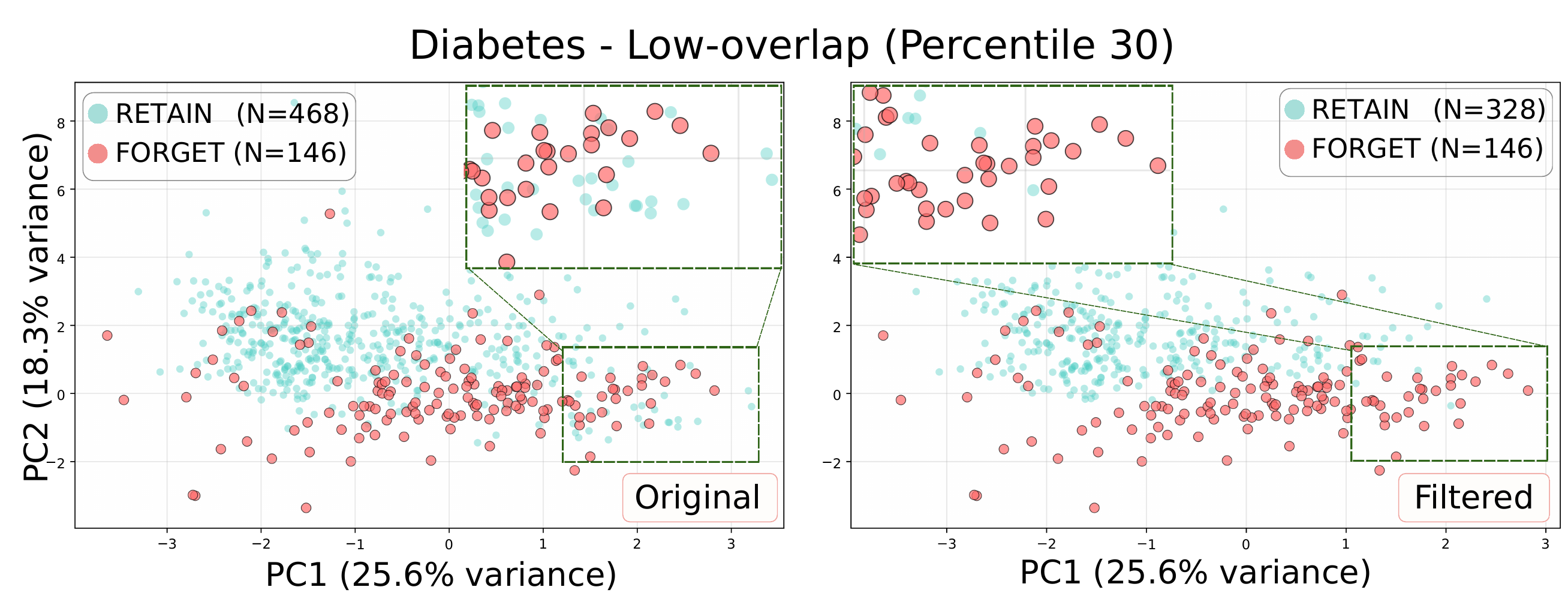}
        \caption{\textsc{Low-overlap}}
        \label{fig:diabetes_change_low}
    \end{subfigure}

    \caption{\small \textbf{PCA visualizations of sample-level unlearning on the Diabetes dataset under high- and low-overlap regimes.}
Each panel compares the representation geometry before (Original) and after conflict-aware retain filtering (Filtered) at percentile 30.
Filtering relaxes retain constraints that lie close to the forget-set distribution, resulting in clearer separation between the forget set and the remaining retain constraints.
    }
    \label{fig:diabetes_change}
\end{figure*}

We evaluate \methodname{} under two complementary unlearning settings that arise naturally in structured tabular data, corresponding to different granularities of deletion. In \textsc{sample-level unlearning}, the deletion target is an entire record, and the goal is to remove the influence of selected rows from the model. In \textsc{feature-level unlearning}, the deletion target is a set of attribute-level evidence spans, and the goal is to suppress the model's reliance on selected columns while preserving overall predictive utility. For fixed-schema tabular data, feature-level unlearning therefore targets selected attribute-name and attribute-value spans, rather than treating every record that contains those columns as a forget sample.

\begin{table*}[t]
\centering
\scriptsize

% ================= LEFT TABLE =================
\begin{minipage}{0.49\textwidth}
\raggedleft
\setlength{\tabcolsep}{2pt}
\renewcommand{\arraystretch}{1.05}

\begin{tabular}{lcccccc}
\toprule
& \multicolumn{3}{c}{\textbf{Qwen2.5 3B}} 
& \multicolumn{3}{c}{\textbf{Qwen2.5 7B}} \\
\cmidrule(lr){2-4} \cmidrule(lr){5-7}
Method
& MIA AUC & $|\Delta|$ & Acc
& MIA AUC & $|\Delta|$ & Acc \\
\midrule
Tuned
& 0.5750 & 0.1520 & 0.8590
& 0.1990 & 0.4980 & 0.8580 \\
\rowcolor{green!15}
Retrained (Oracle)
& 0.4230 & 0 & 0.8160
& 0.6970 & 0 & 0.8220 \\
\midrule
GradDiff
& 0.1080 & 0.3150 & 0.8580
& 0.1010 & 0.5960 & 0.8490 \\
NPO
& 0.0940 & 0.3290 & 0.8580
& 0.0000 & 0.6970 & 0.8580 \\
RMU
& 0.5680 & 0.1450 & 0.8590
& 0.1930 & 0.5040 & 0.8580 \\
SimNPO
& 0.1040 & 0.3190 & 0.8590
& 0.0970 & 0.6000 & 0.8530 \\
\rowcolor{red!15}
\textbf{Ours}
& \textbf{0.3310} & \textbf{0.0920} & 0.8590
& \textbf{0.4560} & \textbf{0.2410} & 0.8580 \\
\bottomrule
\end{tabular}

\vspace{3pt}
% \caption*{\textbf{Dataset 1}}
\end{minipage}
\hspace{0.01\textwidth}
% ================= RIGHT TABLE =================
\begin{minipage}{0.49\textwidth}
\raggedright
\setlength{\tabcolsep}{2pt}
\renewcommand{\arraystretch}{1.05}

\begin{tabular}{lcccccc}
\toprule
& \multicolumn{3}{c}{\textbf{Qwen2.5 3B}} 
& \multicolumn{3}{c}{\textbf{Qwen2.5 7B}} \\
\cmidrule(lr){2-4} \cmidrule(lr){5-7}
Method
& MIA AUC & $|\Delta|$ & Acc
& MIA AUC & $|\Delta|$ & Acc \\
\midrule
Tuned
& 0.7190 & 0.0790 & 0.8590
& 0.6760 & 0.0870 & 0.8580 \\
\rowcolor{green!15}
Retrained (Oracle)
& 0.7980 & 0 & 0.7920
& 0.5890 & 0 & 0.6970 \\
\midrule
GradDiff
& 0.3290 & 0.4690 & 0.2390
& 0.1120 & 0.4770 & 0.0000 \\
NPO
& 0.3150 & 0.4830 & 0.8570
& 0.2450 & 0.3440 & 0.8580 \\
RMU
& 0.7560 & 0.0420 & 0.8580
& 0.5730 & 0.0160 & 0.8570 \\
SimNPO
& 0.6770 & 0.1210 & 0.8410
& 0.1850 & 0.4040 & 0.2390 \\
\rowcolor{red!15}
\textbf{Ours}
& \textbf{0.5130} & \textbf{0.2850} & 0.8600
& \textbf{0.5560} & \textbf{0.0330} & 0.8580 \\
\bottomrule
\end{tabular}

\vspace{3pt}
% \caption*{\textbf{Dataset 2}}
\end{minipage}

\caption{\small \textbf{Sample-level unlearning results on Adult Income under high- and low-overlap regimes.}
Retrained (Oracle) denotes the model trained from scratch only on $\mathcal{D}_r$.
$|\Delta|$ is the absolute difference between each method's loss-based attack AUC and the corresponding Retain Oracle AUC; lower $|\Delta|$ indicates closer agreement with the retain-trained oracle.}
\label{tab:adult_income_sample}

\end{table*}

To study the effect of forget--retain overlap, we construct two forget-set regimes using a k-means partition of the training examples. We first embed each training record with a fixed encoder and run k-means with k = 10 in the embedding space. Each cluster is treated as a candidate forget region, and the remaining training examples define the retain set. We then select \textsc{high-overlap} forget sets from clusters that are less separated from the rest of the training data, and \textsc{low-overlap} forget sets from clusters that are more isolated. After selecting the forget set $\mathcal{D}_f$, the retain set is defined as $\mathcal{D}_r=\mathcal{D}_{\mathrm{train}}\setminus\mathcal{D}_f$. The held-out test set is fixed across overlap regimes and methods. This design isolates the effect of forget--retain overlap while keeping the deletion granularity, model, and evaluation protocol unchanged.

\begin{figure*}[htbp]
    \centering
    \includegraphics[width=\textwidth]{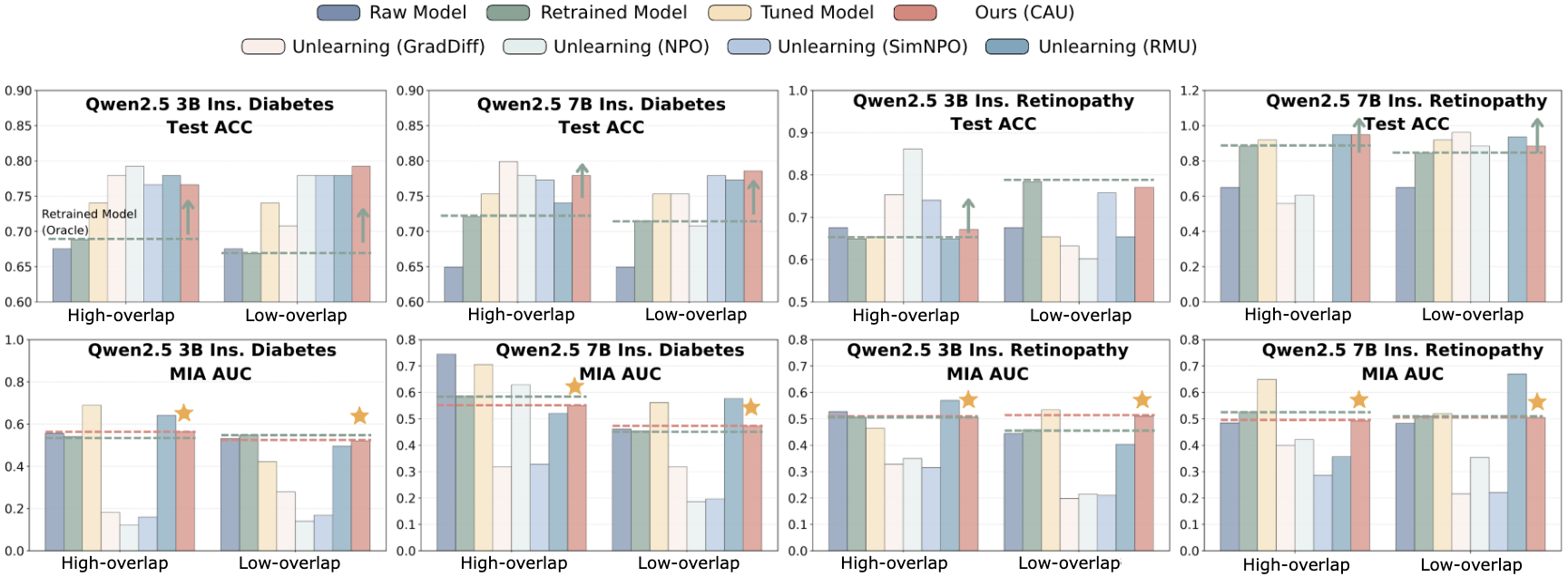}
    \caption{\small \textbf{Sample-level unlearning results on Diabetes and Retinopathy.}
We report test accuracy (top row) and MIA AUC (bottom row) under 
\textsc{high-overlap} and \textsc{low-overlap} forget-set regimes across 
\textsc{Qwen3B} and \textsc{Qwen7B}. 
The retrained model serves as an oracle reference for ideal forgetting 
(green dashed lines). 
Effective unlearning is characterized by privacy behavior close to the 
retrained oracle, while higher utility is better. 
Across settings, \methodname{} achieves more stable forgetting behavior, 
particularly in the \textsc{high-overlap} regime where baseline methods 
exhibit larger variability.
}
    \label{fig:sample_level}
    \vspace{-2mm}
\end{figure*}

\subsection{Experimental Setup}
\textbf{Models and Datasets.}
We evaluate our method using \textsc{Qwen-2.5-3B-Instruct} (\textsc{Qwen3B}) and \textsc{Qwen-2.5-7B-Instruct} (\textsc{Qwen7B}) as backbone language models. Experiments are conducted on three structured tabular prediction tasks: two clinical tasks, Diabetic Retinopathy Debrecen~\cite{b_lint_antal_andr_s_hajdu_2024} and Diabetes, and one non-clinical task, Adult Income~\cite{adult_2}. For each task, we construct forget sets from the cluster-based overlap regimes described above and use the remaining training data as the retain set $\mathcal{D}_r=\mathcal{D}_{\mathrm{train}}\setminus\mathcal{D}_f$.

\noindent\textbf{Baseline Methods.}
We compare \methodname{} against representative unlearning baselines: \textsc{RMU}~\cite{li2024wmdp}, \textsc{NPO}~\cite{zhang2024negative}, \textsc{SimNPO}~\cite{fan2024simplicity}, and \textsc{GradDiff}~\cite{yao2023large}. All implementations are based on Open-Unlearning~\cite{dorna2025openunlearningacceleratingllmunlearning}. We also include a \textsc{Retain Oracle}, trained from scratch only on $\mathcal{D}_r$, as the retraining reference for ideal deletion, and a \textsc{Tuned} model obtained by fine-tuning on the full training set. All unlearning methods are initialized from the tuned checkpoint, which is also reported as a control without unlearning-specific objectives.

\noindent\textbf{Evaluation Metrics.}
Utility is measured by prediction performance on the held-out test set, reported as Test Accuracy. Privacy is assessed using a loss-based membership inference attack AUC. Since the retain-trained model is the oracle reference, we do not treat an AUC of $0.5$ as the universal optimum. Instead, for each dataset, model, and overlap regime, we report the absolute deviation
\[
|\Delta|
=
\left|
\mathrm{AUC}_{\mathrm{method}}
-
\mathrm{AUC}_{\mathrm{retain\ oracle}}
\right|,
\]
where smaller values indicate behavior closer to exact retraining on the retain set. 

\begin{figure}[t!]
    \centering
    \includegraphics[width=0.48\textwidth]{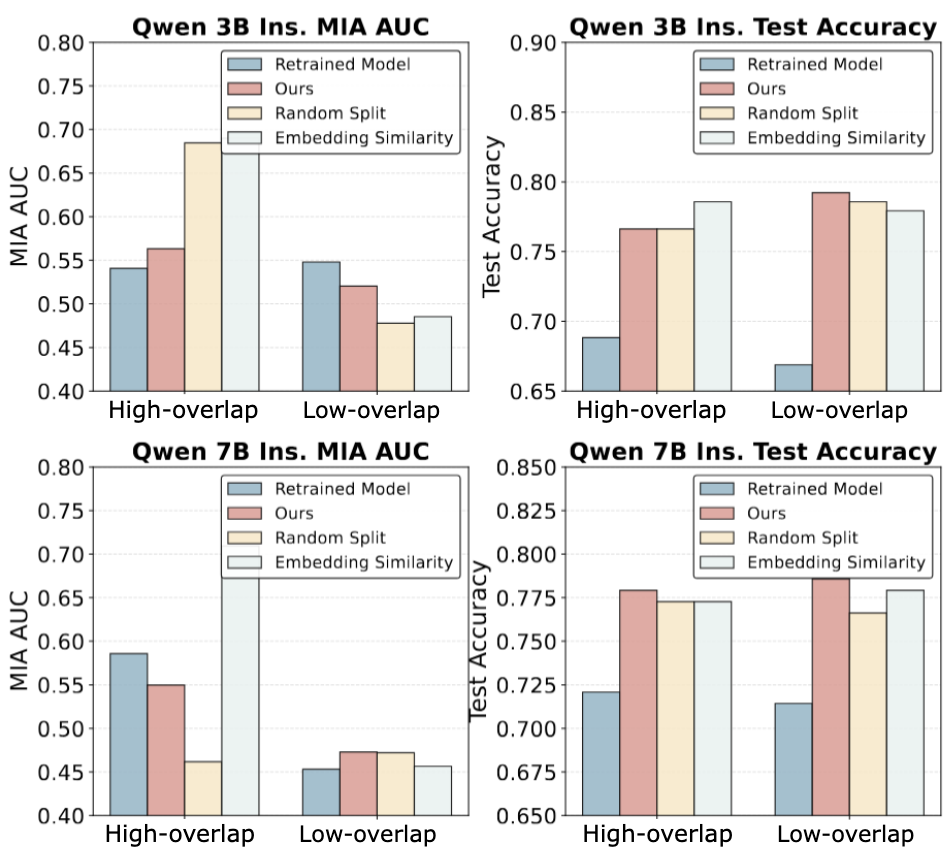}
    \caption{\small \textbf{Ablation study.}
    Comparison of \texttt{CAU} (Ours) with retain selection based on random splitting and embedding similarity, evaluated under two forget-set regimes for Qwen-3B and Qwen-7B.}
    \label{fig:ablation_study}
    \vspace{-3mm}
\end{figure}

\noindent\textbf{Response to Hypothesis: Failure of General Unlearning.}
\label{sec:hypothesis}
We first examine whether standard unlearning methods become unstable 
under strong forget--retain overlap. 
Across sample-level results on clinical tasks and Adult Income 
(Fig.~\ref{fig:sample_level}, Table~\ref{tab:adult_income_sample}), 
baseline methods often deviate substantially from the retain-trained 
oracle in the \textsc{high-overlap} regime while showing inconsistent 
utility preservation. 
The same pattern appears in feature-level unlearning 
(Table~\ref{tab:combined}), where several baselines produce large 
$|\Delta|$ from the oracle or fail to maintain valid predictive behavior.

These results support our hypothesis: when forget and retain examples 
overlap under a shared schema, standard retain-preserving unlearning can 
either over-forget with utility degradation or under-forget relative to 
the retain-trained oracle. 
This instability persists across model sizes and across clinical and 
non-clinical structured tabular tasks.

\begin{mdframed}[userdefinedwidth=.99\linewidth,align=center,
skipabove=3pt,skipbelow=3pt,innerleftmargin=6pt,
innerbottommargin=6pt,innertopmargin=6pt,roundcorner=3pt,
backgroundcolor=cyan!5,linecolor=gray]  
\noindent \textbf{RQ 3}: \textit{Can \methodname{} better approximate 
the retain-trained oracle than general unlearning methods across 
clinical and non-clinical structured tabular tasks?}
\end{mdframed}

\noindent\textbf{Geometric Effects of Conflict-Aware Filtering.}
Fig.~\ref{fig:diabetes_change} visualizes the effect of conflict-aware retain filtering in \methodname{} on sample-level unlearning using PCA projections of the Diabetes dataset. Under both \textsc{high-overlap} and \textsc{low-overlap} regimes, CAU relaxes retain constraints that lie close to the forget-set distribution, leading to clearer separation between the forget set and the remaining retain constraints.

This visualization illustrates the intended effect of conflict-aware filtering: CAU does not remove high-conflict retain samples from evaluation, but reduces their role as optimization constraints during unlearning. This is consistent with the goal of reducing forget--retain interference while preserving retain-region behavior.

\noindent\textbf{Conflict-Aware Unlearning across Sample- and Feature-Level Settings.}
We evaluate \methodname{} under both sample-level and feature-level unlearning settings to assess its effectiveness across different granularities of forgetting. At the sample level, \methodname{} often achieves smaller $|\Delta|$ from the retain-trained oracle while maintaining competitive test accuracy, especially in high-overlap regimes where forget--retain interference is strongest. The Adult Income results further extend the evaluation beyond clinical prediction, showing that the same conflict-aware retain construction can be applied to non-clinical structured tabular data.

At the feature level, \methodname{} achieves reliable oracle agreement under attribute-level forgetting while avoiding large utility degradation in most settings; additional feature-level results on Adult Income are reported in Appendix Tables~\ref{tab:adult_feature_income_3b} and~\ref{tab:adult_feature_income_7b}.Overall, the results suggest that \methodname{} is most useful when forget and retain examples are highly entangled: by relaxing retain constraints near the forget distribution, it reduces optimization conflict and better approximates retain-only retraining. In lower-overlap settings, simpler retain-preserving methods can sometimes match the oracle closely, which is consistent with our hypothesis that CAU is primarily designed for overlap-induced conflict.

\noindent\textbf{Ablation Study.}
We conduct an ablation study to examine the role of retain-constraint construction in our method. Specifically, we compare \methodname{} with two alternatives: random retain selection and embedding-similarity-based selection. As shown in Fig.~\ref{fig:ablation_study}, both ablated variants exhibit less stable oracle agreement, particularly in the \textsc{high-overlap} regime. In contrast, \methodname{} more consistently reduces $|\Delta|$ from the retain-trained oracle while maintaining comparable test accuracy. These results indicate that the benefit of \methodname{} does not come only from reducing the number of retain constraints; it comes from selecting which constraints to relax based on schema-aware proximity to the forget distribution.

\vspace{-2mm}
\section{Conclusion}
\vspace{-2mm}
\label{sec:conclusion}
We study LLM unlearning for serialized tabular data, where shared schemas and similar attribute values can induce strong forget--retain overlap. 
This overlap makes retain preservation an active source of interference: nearby retain samples may constrain the same representations that forgetting needs to change. 
We propose \methodname, a schema-aware retain-construction method that reduces this interference by filtering or relaxing high-conflict retain constraints. 
Across sample-level and feature-level unlearning on clinical and non-medical tabular tasks, \methodname\ better matches a retraining oracle while preserving utility and retain-region behavior. 
These results highlight retain-constraint construction as a key component of reliable tabular LLM unlearning.

\section*{Limitations}
This work focuses on serialized tabular prediction tasks where examples share a fixed schema and exhibit measurable forget--retain overlap. 
\methodname\ may be less beneficial when forget and retain samples are already well separated, or when the tabular schema does not induce meaningful shared attribute-level evidence. 
Because \methodname\ relaxes retain constraints for samples close to the forget distribution, careful retain-region evaluation is necessary to ensure that behavior on high-conflict retain samples is not degraded. 
Future work should study broader tabular schemas, regression tasks, and combinations of conflict-aware retain construction with additional unlearning objectives.

\section*{Acknowledgments}
This research was partially funded by the National Institutes of Health (NIH) under award 1OT2OD038051 and 1R01EB037101-01. The views and conclusions contained in this document are those of the authors and should not be interpreted as representing the official policies, either expressed or implied, of the NIH.
The contributions of Bingqi Shang and Sijia Liu are supported in part by the U.S. National Science Foundation (NSF) under CISE Core Award IIS-2504263 and NSF CAREER Award IIS-2338068, the Coefficient Giving AI Safety Research Award, and the Schmidt Sciences Trustworthy AI Award.

% \bibliography{custom,anthology-overlseaf-1,anthology-overleaf-2}

\bibliography{custom}

\appendix

\section{Appendix}
\label{sec:appendix}
\subsection{Additional Tabular Data Analysis}
\label{app:table_data}
\begin{figure*}[t]
    \centering

    \begin{subfigure}[t]{0.48\textwidth}
        \centering
        \includegraphics[width=\linewidth]{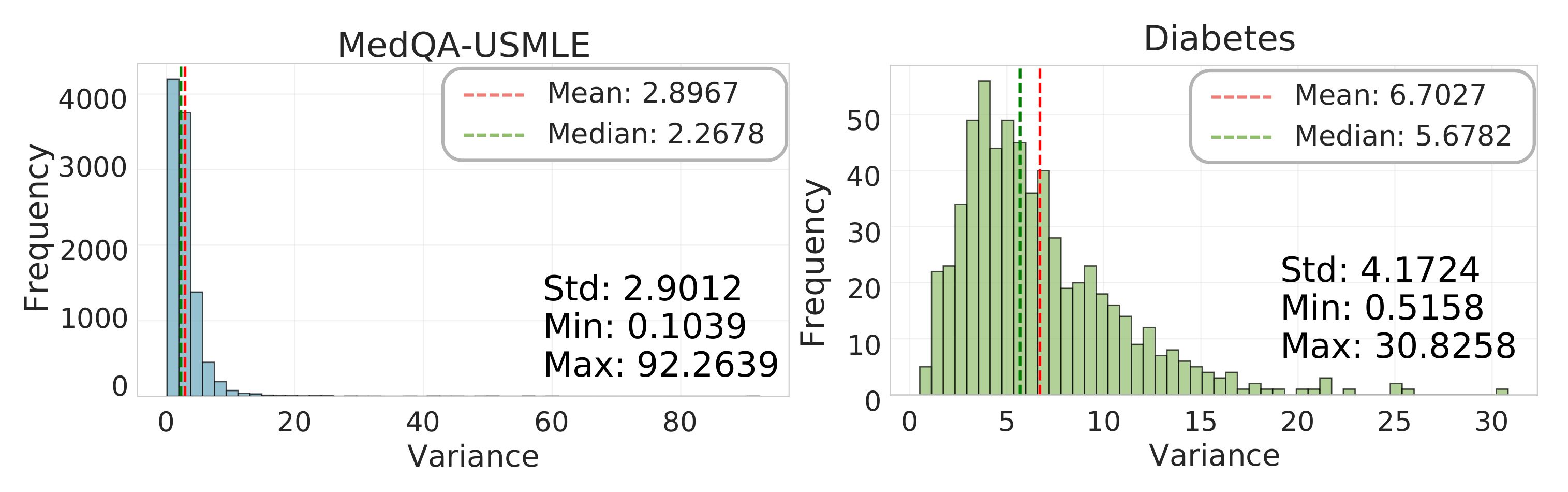}
        \caption{\small Additional Variance Analysis}
        \label{fig:a}
    \end{subfigure}
    \hfill
    \begin{subfigure}[t]{0.48\textwidth}
        \centering
        \includegraphics[width=\linewidth]{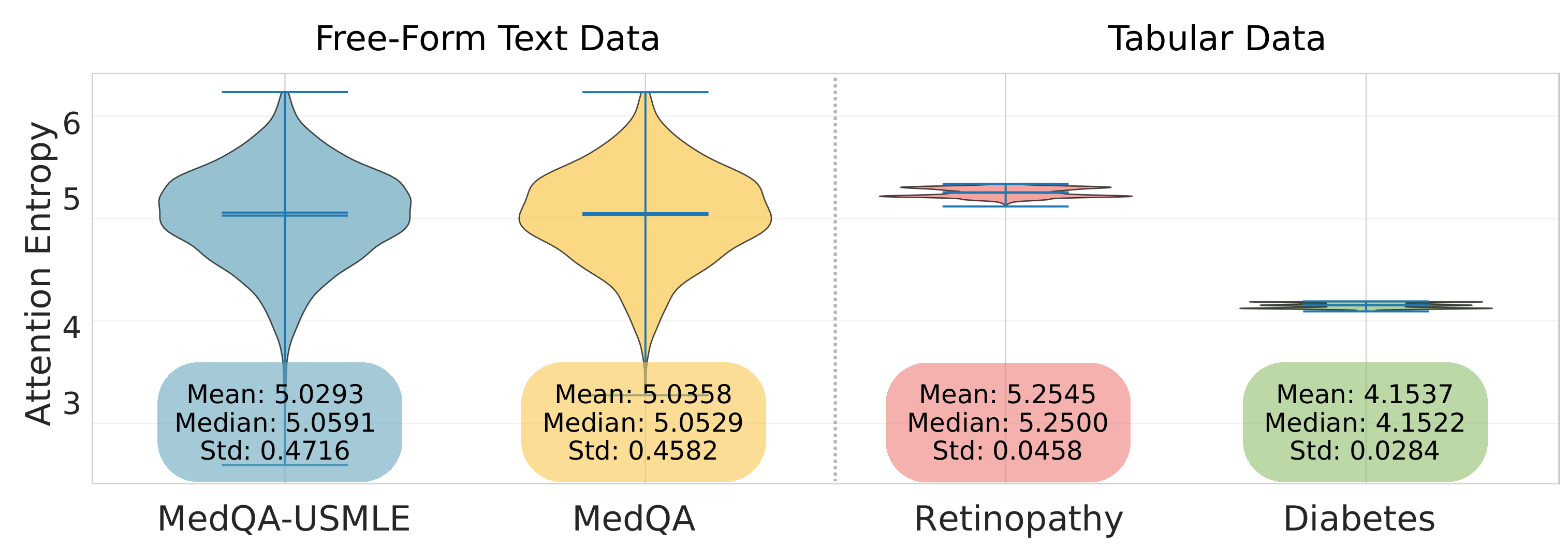}
        \caption{\small Attention Entropy}
        \label{fig:attention_entropy}
    \end{subfigure}

    \caption{\textbf{Token-level sensitivity and attention concentration diagnostics across modalities.} (a) Prediction-logit variance under input-form perturbations for free-form text (MedQA-USMLE) and tabular inputs (Diabetes). Tabular prompts exhibit higher and more heavy-tailed variance, indicating stronger sensitivity to localized input changes.
(b) Token-level attention entropy distributions for free-form and tabular datasets. Lower entropy corresponds to more concentrated attention on a small set of tokens. Tabular inputs often show more concentrated attention patterns, but with dataset-dependent variation. Together, these diagnostics support that structured tabular prompts more frequently induce evidence-concentrated and perturbation-sensitive prediction behavior.
}
    \label{fig:extra_analysis}
\end{figure*}

Fig.~\ref{fig:extra_analysis}(a) shows the distribution of prediction variance under input-form perturbations for free-form text (MedQA-USMLE) and tabular inputs (Diabetes).
For each sample, we generate multiple prompt variants and measure the variance of the model’s target logit across variants.
Higher variance indicates stronger sensitivity to small input-form changes.

We observe a clear difference between the evaluated tabular and free-form text datasets in both scale and spread.
Free-form text exhibits relatively low variance for most samples, with a sharp concentration near small values and only a small number of high-variance outliers.
In contrast, tabular inputs show substantially larger variance and a much heavier-tailed distribution, with both higher mean and wider dispersion.
This suggests that predictions on structured tabular prompts are more sensitive to localized input perturbations.

This behavior is consistent with evidence-concentrated prediction mechanisms: when model decisions depend on a small set of schema-aligned evidence tokens, small perturbations around those tokens can induce disproportionately large output changes.
Such sensitivity may increase the likelihood that unlearning updates applied to shared evidence features also affect nearby retain samples, which is consistent with greater forget–retain interference.

\textbf{Attention patterns are more consistent and constrained in structured tabular inputs.}
Given the last-layer attention matrix $A^{(L)} \in \mathbb{R}^{H \times T \times T}$ extracted from \texttt{Qwen2.5-7B-Instruct}~\cite{qwen2025qwen25technicalreport}, with $H$ heads and sequence length $T$, we compute the attention distribution for each query token $t$ and head $h$:
\begin{equation}
p^{(h)}_{t,j} = \mathrm{softmax}\!\left(A^{(L)}_{h,t,:}\right)_j, \quad j=1,\dots,T.
\end{equation}
We then measure attention entropy:
\begin{equation}
\mathcal{H}^{(h)}(t) = - \sum_{j=1}^{T} p^{(h)}_{t,j} \log p^{(h)}_{t,j},
\end{equation}
and average across heads:
\begin{equation}
\bar{\mathcal{H}}(t) = \frac{1}{H}\sum_{h=1}^{H}\mathcal{H}^{(h)}(t).
\end{equation}
Lower entropy indicates more concentrated attention mass.
As shown in Fig.~\ref{fig:extra_analysis}(b), while the mean attention entropy of tabular inputs is dataset-dependent, tabular data exhibits substantially lower variance in attention entropy across samples.
This indicates that attention patterns induced by structured prompts are more consistent across samples, with the model repeatedly attending to schema-defined token positions.

\paragraph{Summary.}
Across these additional diagnostics, tabular prompts exhibit higher perturbation sensitivity and more regular attention patterns compared to free-form text.
Prediction variance under small input-form changes is larger and more heavy-tailed, and attention distributions are more consistent across samples, indicating repeated use of schema-aligned token positions.
Together, these findings provide additional support for our motivating hypothesis that schema-structured tabular prompts more frequently induce evidence-concentrated and perturbation-sensitive prediction behavior, which may increase forget–retain coupling and make stable unlearning more challenging.

% \subsubsection{Example Data Format and Linearization}
% \label{app:data_format}
\subsubsection{Task Instruction}

\noindent\textbf{Instruction:}
Based on the following medical data, predict the diabetes outcome.
Possible values: 0 (no diabetes) or 1 (has diabetes).
Respond with only `0` or `1`.

\subsubsection{Structured Tabular Format}

\begin{table*}[ht]
\centering
\small
\begin{tabular}{lccccccccc}
\toprule
Sample & Age & BMI & BP & DPF & Glucose & Insulin & Preg. & SkinThick & Outcome \\
\midrule
S1 & 21 & 0.0 & 0 & 0.304 & 84 & 0 & 2 & 0 & 0 \\
S2 & 50 & 28.2 & 82 & 1.282 & 112 & 0 & 9 & 24 & 1 \\
\bottomrule
\end{tabular}
\caption{\small Example samples in structured tabular format. BP = BloodPressure, DPF = DiabetesPedigreeFunction, Preg. = Pregnancies.}
\end{table*}

\subsubsection{Linearized Instruction Format}

\noindent\textbf{Sample S1}
\begin{quote}
Instruction: Based on the following medical data, predict the diabetes outcome. Possible values: 0 (no diabetes) or 1 (has diabetes). Respond with only `0` or `1`.\\
Input: Patient information: Age is 21, BMI is 0.0, BloodPressure is 0, DiabetesPedigreeFunction is 0.304, Glucose is 84, Insulin is 0, Pregnancies is 2, SkinThickness is 0.\\
Output: 0
\end{quote}

\noindent\textbf{Sample S2}
\begin{quote}
Instruction: Based on the following medical data, predict the diabetes outcome. Possible values: 0 (no diabetes) or 1 (has diabetes). Respond with only `0` or `1`.\\
Input: Patient information: Age is 50, BMI is 28.2, BloodPressure is 82, DiabetesPedigreeFunction is 1.282, Glucose is 112, Insulin is 0, Pregnancies is 9, SkinThickness is 24.\\
Output: 1
\end{quote}

\subsection{Implementation and Evaluation Details}
\label{app:impl_eval_details}

\textbf{Baseline Methods.}
We compare \methodname against representative unlearning baselines:
\textsc{RMU}~\cite{li2024wmdp}, which drives forget-sample representations toward a fixed random direction;
\textsc{NPO}~\cite{zhang2024negative} and its stabilized variant \textsc{SimNPO}~\cite{fan2024simplicity}, which suppress the likelihood of forget samples via preference optimization; and
\textsc{GradDiff}~\cite{yao2023large}, which maximizes gradient discrepancies between forget and retain sets. All implementations are based on Open-Unlearning~\cite{dorna2025openunlearningacceleratingllmunlearning}.

\subsubsection{Evaluation Protocol and Fairness Controls}

All unlearning methods are initialized from the same tuned checkpoint trained on the full training set, while the oracle retrained model is trained from scratch using only the retain set. We evaluate two deletion granularities: (i) \textbf{sample-level unlearning}, which removes entire records, and (ii) \textbf{feature-level unlearning}, which suppresses reliance on selected attribute-level evidence (columns). For each setting, we construct HIGH-OVERLAP and LOW-OVERLAP forget sets based on representation-space energy distance, where high-overlap clusters have lower separation from the remaining training data and low-overlap clusters are more isolated.

We report two metrics throughout: (1) test accuracy on a held-out test set to measure utility, and (2) loss-based membership inference attack (MIA) AUC to measure privacy behavior. Effective unlearning is measured by closeness to the retain-trained oracle, using $|\Delta| = \left|\mathrm{AUC}_{\text{method}} - \mathrm{AUC}_{\text{retain oracle}}\right|$.

All methods use identical data splits, base checkpoints, and evaluation pipelines for fair comparison.

\subsubsection{Feature-Level Unlearning Setup}

Let $A$ denote the full set of tabular attributes and $A_f \subset A$ the target feature subset to be forgotten. Each tabular record is serialized into a structured text prompt with explicit feature-name and feature-value tokens. We define feature-level forgetting with respect to token spans corresponding to attributes in $A_f$.

Specifically, for each serialized sample, we identify token positions associated with:
(i) the feature name, and
(ii) its value field.
These positions are marked as \textit{feature evidence tokens}. During unlearning, forgetting-oriented objectives are applied only to these marked tokens, while retain-preservation losses are computed on the full sequence.

The forget set for feature-level unlearning is the selected high-overlap or low-overlap row group; within each forget sample, the objective targets only token spans associated with $A_f$. The retain set is $D_r = D_{\mathrm{train}} \setminus D_f$. Test-time evaluation is performed on the standard held-out test split without masking, so utility reflects end-task performance under normal inputs.

% \subsubsection{CAU Filtering and Distance Computation Details}

% For conflict-aware unlearning, we compute the Forget-Normalized Distance (FND) between each retain sample representation and the forget-set representation distribution. Representations are extracted from the final hidden layer using mean pooling over tokens. Per-dimension normalization uses forget-set mean and standard deviation with a small constant $\epsilon$ for numerical stability.

% Retain samples are ranked by FND, and only the lowest-conflict subset is used in the retain-preservation objective. We select retain samples using a percentile threshold $p$ (default $p=30$). We reuse the same frozen encoder and representation extraction pipeline across all methods.

\subsubsection{Unlearning Objective and Hyperparameters}

All unlearning methods are run with matched optimization budgets. We use the same optimizer, learning rate, batch size, and number of update steps across CAU and all baselines.

\begin{table}[h]
\centering
\small
\begin{tabular}{l l}
\toprule
Optimizer & AdamW \\
Learning rate & 1e-5 \\
Batch size & 4 \\
Update steps & 38 (1 epoch) \\
Weight decay & 0.01 \\
\midrule
CAU percentile $p$ & 30 \\
RMU strength $c$ & 20 \\
Loss mixing weight $\alpha$ & 1.0 \\
Representation layer $\ell$ & 7 \\
\bottomrule
\end{tabular}
\caption{\small Shared optimization settings and key unlearning hyperparameters used in all experiments.}
\label{tab:unlearning_hparams}
\end{table}
The representation-level unlearning loss (e.g., RMU-style objectives) is applied at layer $\ell$ using hidden states at the marked token positions (feature evidence tokens for feature-level unlearning, full sequence for sample-level unlearning unless otherwise stated).

\subsubsection{Membership Inference Attack Protocol}

We adopt a loss-based membership inference attack. For each example, we compute a scalar score equal to the average per-token negative log-likelihood (NLL) of the ground-truth output under teacher forcing.

Forget-set training samples are treated as members. An equal-sized set of non-members is drawn from a disjoint held-out split that is never used for training or unlearning. We compute ROC curves using these scores and report the corresponding AUC. All reported MIA AUC values use a fixed evaluation seed and identical member/non-member sampling sizes across methods.

\subsubsection{Reproducibility Notes}

All experiments use fixed data splits, fixed serialization templates, and fixed token-span rules for feature evidence identification. Checkpoint selection, evaluation scripts, and MIA scoring are shared across methods to avoid evaluation bias. Full configs and scripts will be released with the code.

\subsection{Additional Adult Income Feature-Level Results}

\begin{table}[t]
\centering
\small
\begin{tabular}{lccc}
\toprule
Method & Test Acc & MIA AUC & $|\Delta|$ \\
\midrule
Tuned & 0.8591 & 0.7194 & 0.0785 \\
Retrained (Oracle) & 0.7916 & 0.7979 & 0.0000 \\
GradDiff & 0.2394 & 0.3295 & 0.4684 \\
NPO & 0.8567 & 0.3147 & 0.4832 \\
RMU & 0.8597 & 0.6765 & 0.1214 \\
SimNPO & 0.8411 & 0.5130 & 0.2849 \\
Ours & 0.8582 & 0.7561 & 0.0418 \\
\bottomrule
\end{tabular}
\caption{Feature-level unlearning results on Adult Income with Qwen2.5-3B-Instruct. MIA AUC is the loss-based membership inference attack AUC, and $|\Delta|$ is computed relative to the retrained oracle.}
\label{tab:adult_feature_income_3b}
\end{table}

\begin{table}[t]
\centering
\small
\begin{tabular}{lccc}
\toprule
Method & Test Acc & MIA AUC & $|\Delta|$ \\
\midrule
Tuned & 0.8577 & 0.6760 & 0.0869 \\
Retrained (Oracle) & 0.6970 & 0.5891 & 0.0000 \\
GradDiff & 0.0000 & 0.1118 & 0.4773 \\
NPO & 0.8577 & 0.2451 & 0.3440 \\
RMU & 0.8570 & 0.5556 & 0.0335 \\
SimNPO & 0.2394 & 0.1848 & 0.4043 \\
Ours & 0.8579 & 0.5734 & 0.0157 \\
\bottomrule
\end{tabular}
\caption{Feature-level unlearning results on Adult Income with Qwen2.5-7B-Instruct. MIA AUC is the loss-based membership inference attack AUC, and $|\Delta|$ is computed relative to the retrained oracle.}
\label{tab:adult_feature_income_7b}
\end{table}

\subsection{Additional Related Work}
\label{sec:related_work_appendix}

\textbf{Machine Unlearning.}
Machine unlearning aims to remove the influence of a designated forget set from a trained model while preserving performance on the remaining retain data, offering an efficient alternative to full retraining under deletion requests~\cite{jang2023knowledge,yao2023large,liu2025rethinking}.
Early work explored training-time mechanisms such as data sharding and slicing to enable efficient deletion~\cite{bourtoule2020machineunlearning,ginart2019makingaiforgetyou,graves2020amnesiacmachinelearning,cao2015towards}.
Subsequent approaches focus on post-hoc updates that approximate retraining using fine-tuning, gradient-based corrections, distillation, or regularization objectives~\cite{golatkar2020eternalsunshinespotlessnet,golatkar2020forgettingoutsideboxscrubbing,guo2023certifieddataremovalmachine,ding2025understandingfinetuningapproximateunlearning,neel2020descenttodeletegradientbasedmethodsmachine}.

Recent work has extended unlearning to large-scale and pre-trained language models, systematically evaluating unlearning strategies and their trade-offs between privacy and utility~\cite{yao2024machineunlearningpretrainedlarge,gundavarapu2024machineunlearninglargelanguage,eldan2023whosharrypotterapproximate,peng2025forgetknowrememberuse}. These studies highlight both the feasibility and instability of post-hoc unlearning in LLMs, motivating more structured and conflict-aware update rules.

More recently, geometry-aware and bilevel optimization formulations have been proposed to explicitly control forget–retain interference during unlearning, including projection-based, subspace-restricted, and conflict-aware optimization strategies (e.g., SFR-on, BLUR)~\cite{huang2024unifiedgradientbasedmachineunlearning,reisizadeh2025blurbileveloptimizationapproach,hu2025blurbenchmarkllmunlearning}. These methods model unlearning as a constrained or multi-objective optimization problem and aim to reduce gradient or representation-level conflicts between forget and retain objectives. Our work is complementary but distinct: instead of modifying the optimization solver, we introduce a data-level conflict filtering mechanism tailored to structured tabular prompts, which can be combined with existing unlearning objectives such as RMU~\cite{li2024wmdpbenchmarkmeasuringreducing}.

\textbf{Unlearning under Data Correlation and Privacy Risks.}
Recent studies highlight that unlearning becomes substantially more challenging when the forget set is statistically or representationally entangled with retain data, such as overlapping feature distributions or shared internal representations~\cite{zhao2024makes}.
Under these conditions, unlearning updates intended to remove specific samples may propagate to nearby retain points, causing unintended utility degradation and unstable forget-utility trade-offs~\cite{zhao2024makes,golatkar2020eternalsunshinespotlessnet,golatkar2020forgettingoutsideboxscrubbing,ding2025understandingfinetuningapproximateunlearning}.

Several recent methods explicitly model and manage forget–retain conflicts through geometry-aware or bilevel formulations, including subspace filtering, projection-based retention, and constrained optimization strategies (e.g., SFR-on, BLUR)~\cite{huang2024unifiedgradientbasedmachineunlearning,reisizadeh2025blurbileveloptimizationapproach,hu2025blurbenchmarkllmunlearning,neel2020descenttodeletegradientbasedmethodsmachine,golatkar2020eternalsunshinespotlessnet}, which aim to decouple forget and retain gradients or representations. Related lines of work also explore conflict-guided localization and selective update regions for LLM unlearning (e.g., CLUE)~\cite{chen2025clueconflictguidedlocalizationllm}, identifying parameters or components most responsible for forget-set behavior. Distribution-level preference-based unlearning methods, such as DiPO~\cite{qin2025distributionpreferenceoptimizationfinegrained}, further reframe forgetting as a preference or distribution-matching problem, optimizing relative likelihood constraints between forget and retain outputs rather than pointwise losses. Compared to NPO- or SimNPO-style objectives~\cite{zhang2024negative,fan2024simplicity}, these approaches provide a softer, distributional control over trade-offs. Our approach is aligned with this conflict-aware perspective but differs in mechanism and scope: instead of modifying the optimizer or parameter subspace, we target the retain set itself and introduce a modality-aware retain filtering rule that reduces conflict exposure before optimization, which is particularly effective for schema-structured tabular inputs.

Moreover, the unlearning process itself may introduce additional privacy vulnerabilities, as reconstruction-style attacks can recover deleted training samples from unlearning outputs or update signals~\cite{bertran2024reconstructionattacksmachineunlearning,shang2025forgetting}. These findings motivate selective, structure-aware, and correlation-sensitive unlearning strategies that explicitly account for data overlap and representation coupling~\cite{neel2020descenttodeletegradientbasedmethodsmachine,guo2023certifieddataremovalmachine,li2025loreundataimplicitlyprovides}.

\subsection{LLM Usage}
Large Language Models (LLMs) were used to assist with editing and polishing the writing of this manuscript. Specifically, the LLM was used to refine wording, improve clarity, and enhance overall readability.

The LLM was not involved in the development of research ideas, methodology, data analysis, or experimental design. All technical content, interpretations, and conclusions were independently developed and validated by the author.

The author takes full responsibility for the content of the manuscript, including any text that was edited with LLM assistance. All suggested revisions were carefully reviewed for accuracy and originality.

\end{document}